\documentclass[conference]{IEEEtran}
\usepackage{newtxtext,newtxmath}
\usepackage{graphicx,booktabs,amsmath,url,xcolor,tikz,pgfplots,upquote}
\pgfplotsset{compat=1.18}
\usetikzlibrary{positioning,arrows.meta,fit}

\begin{document}

\title{Hybrid Product Search Without the Copy: Lexical, Semantic and Filtered Retrieval Served Directly from a Data Lake}

\author{
\IEEEauthorblockN{Raghvendra Yadav, Xiang Fu, Chinmay Soman}
\IEEEauthorblockA{StarTree Inc., Mountain View, CA, USA\\ \{raghvendra, xiangfu, chinmay\}@startree.ai}
}

\maketitle

\begin{abstract}
E-commerce catalogs increasingly live in lakehouse tables: Apache Iceberg metadata over Parquet files in object
storage. Product search, however, still runs on copies of that catalog, in a lexical engine and a vector database,
each with its own ingestion pipeline, schema, staleness window and bill. We show that search applications can be
hosted directly on the data lake, with the lakehouse table as the single system of record. We extend Apache Pinot, a
real-time OLAP engine that already serves Iceberg tables in place, with BM25 relevance scoring over its Lucene text
index, corpus-wide term statistics so that scores are comparable across files, and reciprocal rank fusion in plain
SQL; a real-time table fed by a stream can sit behind the same table, merged by the broker. On the Amazon
ESCI benchmark (1.2M US products, 22,458 test queries), hybrid search served from the lake matches exact retrieval
(nDCG@10 .361). The lake table is queryable 149\,s after its files are registered, and a lake commit is searchable
in 3.8\,s, close to indexing directly into a search engine. On a brand-sorted lake of 277M products (0.8\,TB), one
16-vCPU node answers brand-scoped vector, BM25 and client-fused hybrid searches in 6.5--9\,ms (p50), sustains 1,100 vector and 635
BM25 searches per second at under 180\,ms p99, and makes a commit searchable in 4.3\,s. Search adds only its indexes: no copy of
the catalog, no sync pipeline and no second schema, which cuts storage, rebuild time and the operational work of
keeping copies consistent.
\end{abstract}


\section{Introduction}
Product catalogs are now routinely stored as lakehouse tables~\cite{lakehouse}: Parquet files in object storage
described by an Iceberg~\cite{iceberg} snapshot. Analytics, feature pipelines and reporting read those files in place.
Search does not. A typical retail search stack copies the catalog into a lexical engine for BM25~\cite{bm25} and into
a vector database for dense retrieval, then fuses the two result lists in the application. Each copy has its own
ingestion job, schema mapping, access control and freshness lag. Search freshness is then set by the slowest of these
pipelines, every schema change has to be made three times, and every backfill, re-index and consistency check is
operational work that exists only because the data was copied.

We ask whether search applications---lexical, semantic and filtered product search---can run directly on the
lake's own files, with the lake as the only system of record and no loss of relevance, freshness or interactive latency. We build on Apache Pinot~\cite{pinot}, which can already serve
Iceberg tables in place by treating each Parquet data file as a segment, and make three contributions:
\begin{enumerate}
  \item \textbf{Relevance over files we do not own.} We add BM25 scoring to Pinot's Lucene text index
  (\texttt{TEXT\_SCORE}), with a two-phase protocol that gathers corpus-wide term statistics
  (\texttt{TEXT\_SCORE\_STATS}), in the spirit of Elasticsearch's DFS phase, so scores are comparable across files.
  \item \textbf{Hybrid retrieval in SQL.} BM25, approximate nearest-neighbour (ANN) search with structured filter
  pushdown, and reciprocal rank fusion (RRF)~\cite{rrf} compose in one SQL query; corpus-wide BM25 statistics take one
  preceding metadata-only request.
  \item \textbf{An evaluation on Amazon ESCI}~\cite{esci}: relevance on all 22,458 test queries with paired
  significance tests, onboarding, freshness, latency, throughput on a 277M-product lake, and infrastructure cost.
\end{enumerate}
Quality served from the files matches exact retrieval; the table is queryable minutes after its files are
registered and a lake commit is searchable in seconds; selective queries answer in milliseconds; and the search tier
stores only its indexes, never the rows. Selective queries, the common case in product search, are where the design is
strongest: file statistics skip most of the lake before any index is touched.

\section{Background and Related Work}
\textbf{Lakehouse tables.} Iceberg tracks a table as a sequence of immutable snapshots, each listing Parquet data
files and (in format v3) deletion vectors. Engines read the files with range requests against object storage.

\textbf{Lexical and dense retrieval.} BM25~\cite{bm25} scores a document by term frequency, inverse document
frequency (IDF) and length normalisation; the IDF and average length are corpus statistics, so a sharded index
returns comparable scores only if shards share them (Elasticsearch's DFS phase~\cite{elasticsearch}). Dense retrieval embeds queries and
products with a bi-encoder and searches an ANN index such as HNSW~\cite{hnsw}. Dense retrievers~\cite{dpr} are brittle out of domain, where BM25 remains a hard
baseline~\cite{beir}; on product search the two are complementary~\cite{esci}, and filters can be applied before, during
or after the graph search, with filter-aware graph indexes applying them during traversal~\cite{filtereddiskann,acorn}. Fusion can be score-based (a convex combination of normalised scores) or rank-based;
RRF~\cite{rrf} needs no score calibration and is supported natively by most engines, although tuned convex combinations can
do better with in-domain labels~\cite{fusion}.

\textbf{Hybrid engines and search near the lake.} Vespa~\cite{vespa}, Elasticsearch~\cite{elasticsearch} and OpenSearch combine lexical,
vector and filtered retrieval with built-in fusion, but over their own ingested copy. Vector stores such as Milvus~\cite{milvus}, Qdrant and Weaviate
add hybrid and filtered search; Lance~\cite{lance} and turbopuffer~\cite{turbopuffer} keep their own index formats on object storage, and lakehouse
vendors offer managed vector search synced from tables~\cite{dbvs}. Lakehouse formats such as Delta Lake~\cite{deltalake} and Iceberg made the
lake transactional, which is what lets a search tier bind its indexes to a consistent snapshot. Closest to our work,
Rottnest~\cite{rottnest} builds substring, BM25 and vector indexes next to Parquet files on object storage, an index
lake for Anserini~\cite{geese} keeps postings as lake tables, and per-file ANN indexes composed with Iceberg file
pruning serve filtered vector search~\cite{lakepann}. We add corpus-wide BM25 statistics across per-file indexes,
index lifecycle bound to Iceberg snapshots, one table across lake and stream, and a product-search evaluation. Our setting differs
from all of the above: the source files belong to the lake, are written by other engines and change by snapshot; the
search engine adds indexes on its own tier but never ingests or rewrites the rows.

\section{Architecture}
\begin{figure*}[t]
  \centering
  \begin{tikzpicture}[font=\small, node distance=6mm and 10mm,
      box/.style={draw, rounded corners=2pt, align=center, inner sep=3pt, minimum height=9mm},
      lake/.style={box, fill=blue!7}, srv/.style={box, fill=orange!10}, brk/.style={box, fill=green!8},
      ar/.style={-{Latex[length=1.6mm]}}]
    \node[lake, text width=34mm] (cat) {Iceberg REST catalog\\snapshot $s$ (data files, DVs)};
    \node[lake, text width=40mm, right=34mm of cat] (pq) {Parquet data files\\(object storage, owned by the lake)};
    \node[box, text width=34mm, below=9mm of cat] (ctl) {Controller sync task\\1 data file $=$ 1 segment};
    \node[srv, text width=40mm, below=9mm of pq] (idx) {Servers: Lucene text index +\\HNSW index, built at load};
    \node[srv, text width=28mm, right=10mm of idx] (pc) {footer cache +\\page cache (prefetch)};
    \node[brk, text width=150mm, below=8mm of idx, xshift=-12mm] (q)
      {Broker, pinned to snapshot $s$:\quad (1) \texttt{TEXT\_SCORE\_STATS} $\rightarrow$ corpus-wide BM25 statistics
       \quad (2) BM25 $\parallel$ ANN (+ filter pushdown) per segment $\rightarrow$ RRF in SQL $\rightarrow$ top-$k$};
    \draw[ar] (cat) -- node[left]{new snapshot} (ctl);
    \draw[ar] (ctl) -- node[above]{assign segments} (idx);
    \draw[ar, dashed] (pq) -- node[right]{range reads} (idx);
    \draw[ar, dashed] (pq.east) -| (pc.north);
    \draw[ar] (q.north -| idx.south) -- (idx.south);
  \end{tikzpicture}
  \caption{Hybrid retrieval served from lakehouse files. Rows stay in the lake (blue); only indexes and caches live on
  the search tier (orange). Queries are routed to exactly one published snapshot.}
  \label{fig:arch}
\end{figure*}
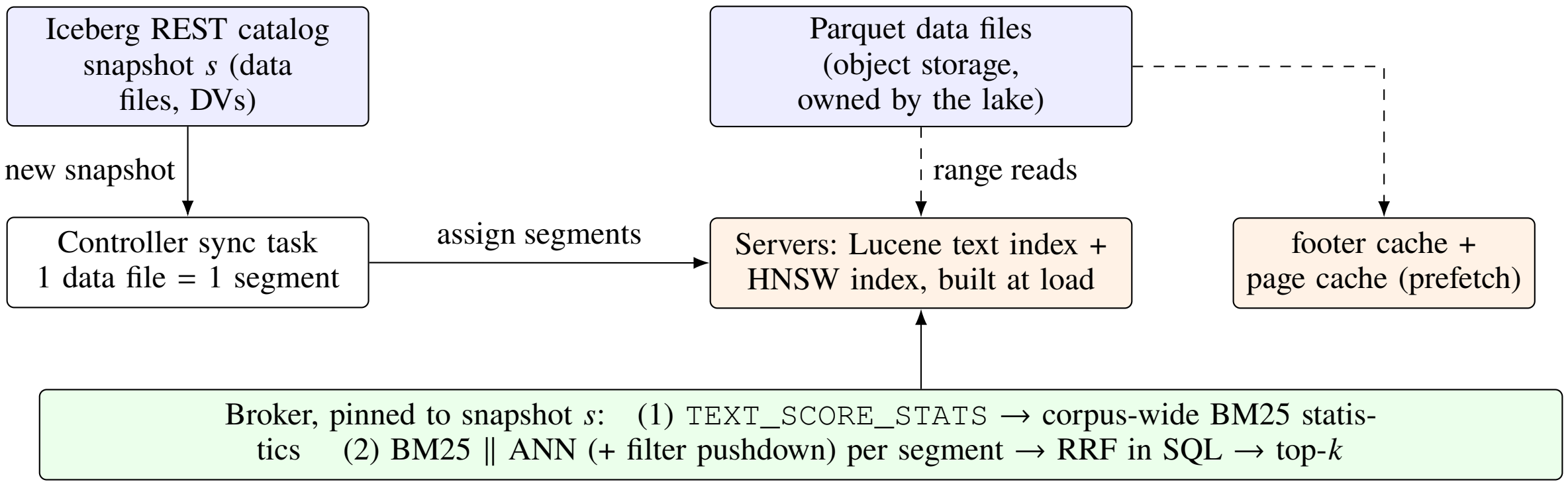

\subsection{Indexes over files we do not own}
An external table maps every Parquet data file in the current Iceberg snapshot to one Pinot segment. Registration
copies no rows: the segment directory holds metadata and a pointer to the source file, and the source file
\emph{is} the forward index. Secondary indexes---here a Lucene text index on the product text and an HNSW index on
the embedding column---are built by the server when the segment loads, from column chunks read with range requests,
and stored on the server's tier next to the segment metadata; nothing is written into the lake. The indexes are
the only search-side state. On ESCI the
HNSW index takes about 3.1\,KB per product and the text index about 222\,B, so vector indexes dominate index storage (Section~\ref{sec:cost}).

\subsection{Snapshot-consistent freshness}
A controller task follows the Iceberg catalog. New data files become new segments; when all segments of a snapshot
are loaded, the snapshot is published and the broker routes queries to exactly that snapshot's segments, pinning one
snapshot per query. Removed files leave the routing table only when the snapshot that dropped them is published, so
readers never see a half-applied commit. Iceberg v3 deletion vectors are applied at query time.

\subsection{Building indexes for a new file}\label{sec:build}
A new data file becomes queryable once its own indexes exist, so the build sits on the freshness path. Three choices
keep it short. The HNSW writer buffers the whole file (512\,MB) so that its graph is built once rather than flushed in
small pieces and merged; small files, typical of a streaming commit, are built from several threads, each into its own
Lucene segment, and attached without a rebuild; and the controller polls segment state every 500\,ms. Nothing is
force-merged afterwards: each file keeps its own graph for its whole life.

\subsection{Cold and warm read paths}
Footers are read once and cached; later reads fetch single pages through the Parquet page index and a local page
cache. Index files are persisted next to the segment metadata on the server's tier and memory-mapped, so index lookups
never touch object storage and the lake serves only the columns a query returns. Lucene reads the index buffers
directly, without copying each page into a heap array (Section~\ref{sec:insights}).

\subsection{Streaming and batch in one table}\label{sec:kappa}
New and changed products can also arrive on a Kafka stream into a Pinot real-time table, while the catalog of record
stays in Iceberg as an external table. Both sit behind one hybrid or logical table: the broker keeps a time boundary
between them, routes each query to lake segments for data before the boundary and to real-time segments for data
after it, and merges the results, so applications see a single table. Search and analytics run over both through the
same SQL, and data lands once in the lake: one pipeline and one query surface, a kappa-style design rather than
separate batch and speed layers. This paper evaluates the lake path; the streaming path is Pinot's existing real-time
ingestion and is not measured here.

\subsection{BM25 scoring with corpus-wide statistics}
Pinot's text predicate \texttt{TEXT\_MATCH} was a boolean filter: its Lucene collectors ran without scoring. We add
\texttt{TEXT\_SCORE(column, query)}, which parses the query exactly as \texttt{TEXT\_MATCH} does and returns the
Lucene BM25 score ($k_1{=}1.2$, $b{=}0.75$), or 0 for non-matching documents. Per segment it runs one scored search on
the first block and serves later blocks from a document-to-score map. No re-indexing is needed: term frequencies and
length norms are already stored.

For a query $q$ and document $d$,
\begin{equation}\small \mathrm{BM25}(q,d)=\sum_{t\in q}\ln\!\Big(1+\frac{N-n_t+0.5}{n_t+0.5}\Big)\,
\frac{f_{t,d}\,(k_1+1)}{f_{t,d}+k_1\big(1-b+b\,|d|/\overline{|d|}\big)},
\label{eq:bm25}
\end{equation}
where $f_{t,d}$ is the term frequency, $N$ the number of documents, $n_t$ the number containing $t$ and
$\overline{|d|}$ the mean document length. $N$, $n_t$ and $\overline{|d|}$ are \emph{corpus} statistics, and a
Lucene index computes them over itself. Because one file is one segment, per-segment IDF would make scores
incomparable across files: a term that is rare in one file is common in another, and each file's own statistics
leak into the global top-$k$. We therefore add a
statistics phase, \texttt{TEXT\_SCORE\_STATS(column, query)}, an aggregation that reads only index metadata and
returns, for each query term, its document frequency and total term frequency, plus document and token counts; the
broker sums them across segments and servers. Passing the merged statistics to \texttt{TEXT\_SCORE} as a fourth argument substitutes the global $N$, $n_t$ and
$\overline{|d|}$ in Eq.~\eqref{eq:bm25}, so every segment scores against corpus-wide IDF and average length. The
statistics query touches only the term dictionary and per-field counts; it reads no Parquet data. Section~\ref{sec:abl} measures the effect on ESCI. The statistics ignore row
filters and can drift if the snapshot changes between the two requests, as with any DFS-style protocol.

\subsection{Hybrid query execution}
Vector search uses \texttt{VECTOR\_SIMILARITY} on the HNSW index. Structured filters are pushed down; per segment a
selectivity heuristic decides whether the filter bitmap restricts the graph search (selective filters), is applied to
the ANN candidates (broad filters), or the segment is scanned exactly (very few candidates). Fusion
needs no new operator. RRF scores a document by its ranks $r_i(d)$ in the fused lists,
\begin{equation}
\mathrm{RRF}(d)=\sum_{i}\frac{1}{k+r_i(d)},\qquad k=60,
\label{eq:rrf}
\end{equation}
and in Pinot's multi-stage engine this is plain SQL over two ranked candidate lists, with a deterministic
tie-breaker on the product id:
\begin{footnotesize}
\begin{verbatim}
WITH t AS (SELECT id, ROW_NUMBER() OVER
    (ORDER BY s DESC, id) r FROM (SELECT id,
    TEXT_SCORE(text,:q,'',:stats) s
    FROM products WHERE TEXT_MATCH(text,:q)
    ORDER BY s DESC LIMIT 100)),
 v AS (SELECT id, ROW_NUMBER() OVER
    (ORDER BY d, id) r FROM (SELECT id,
    cosineDistance(emb,:qv) d FROM products
    WHERE VECTOR_SIMILARITY(emb,:qv,100)
    ORDER BY d LIMIT 100))
SELECT id, SUM(1.0/(60+r)) rrf FROM
  (SELECT * FROM t UNION ALL SELECT * FROM v)
GROUP BY id ORDER BY rrf DESC LIMIT 10
\end{verbatim}
\end{footnotesize}
Structured predicates are added to both branches and pushed down to the segment scans.

\section{Experimental Setup}\label{sec:setup}
We ask four questions: does serving from the lake cost relevance (Section~\ref{sec:quality}); how fast and how much
does it serve at scale (Section~\ref{sec:scale}); what does it cost to onboard, store and run
(Sections~\ref{sec:cost}--\ref{sec:infra}); and how fresh is it (Section~\ref{sec:fresh}).

\textbf{Data.} Amazon ESCI~\cite{esci}, US locale: 1,215,854 products and the official test split of 22,458 queries
with 425,762 judgments. Gains for nDCG are E=1, S=0.1, C=0.01, I=0; Recall@100 counts Exact products. Unjudged
products have zero gain, so absolute scores are low and the comparison between rows is what matters. The indexed text
is the title, brand, colour and the first 600/400 characters of bullets and description.

\textbf{Embeddings.} BGE-base-en-v1.5~\cite{bge} (768 dimensions, L2-normalised), computed once and stored as a Parquet
column; every system uses the same vectors and the model's query prefix.

\textbf{Lake.} One Iceberg table (format v2, REST catalog) in Amazon S3, partitioned by
\texttt{locale} (identity transform; all US products fall in one partition) and written by pyiceberg as 33 Parquet
files of about 133\,MB (4.38\,GB): one row group per file, zstd compression, 64\,KB data pages and a page index
(ColumnIndex and OffsetIndex) on every column chunk, rows in the dataset's original order. Relevance (Table~\ref{tab:quality}) was computed over the same files and build on a development
machine; a 2,000-query sample re-run on EC2 matched to four decimals. Performance runs on AWS (us-west-2) with the
lake in S3 in the same region: onboarding, freshness and the engine baselines on r7g.2xlarge nodes (8 Graviton3 vCPUs, 64\,GB), and latency
and throughput on one i4i.4xlarge (16 vCPUs, 128\,GB, local NVMe) with a separate load generator in the same zone, one
process per client, at 1 to 128 clients over 1,000 sampled queries after 200 warm-up queries. 

\textbf{Filtered hybrid.} Both retrievers and the exact reference are restricted to branded products (94.1\%).

\textbf{Systems.} (i) \emph{Pinot lake-native}: external table over the Iceberg files, Lucene text index and HNSW
vector index built over the remote files; (ii) Elasticsearch 9.5.5 and (iii) OpenSearch 3.9.0 with BM25, float32 HNSW ($M{=}16$,
$\mathit{ef}_{c}{=}100$, 1{,}000 candidates) and their native RRF, following each vendor's guidance for a
read-heavy single node: one shard and no replicas, a 16\,GB heap with the rest of memory left to the file
cache, the index force-merged to one segment after loading, and \texttt{\_source} disabled. Every system uses the same analyzer (standard tokenizer,
lowercase, English stop words, no stemming). (iv) \emph{Reference}: exact BM25 (bm25s, same analyzer) and exact inner-product search.

\textbf{Metrics.} nDCG@10 and Recall@100 per query, compared with a paired two-sided randomisation test (10,000 sign
flips); client-side latency (p50, p99) and throughput, at 32 clients as the median of repeated passes; time from registration or commit
to fully searchable; and bytes stored beyond the lake.

\section{Results}\label{sec:results}
\subsection{Retrieval quality}\label{sec:quality}
\begin{table*}[t]
  \centering
  \caption{Retrieval quality on the ESCI-US test split (22,458 queries). Filtered hybrid restricts both retrievers to
  branded products.}
  \label{tab:quality}
  \setlength{\tabcolsep}{6pt}
  \begin{tabular}{lcccccccc}
    \toprule
    & \multicolumn{2}{c}{BM25} & \multicolumn{2}{c}{Vector (HNSW)} & \multicolumn{2}{c}{Hybrid RRF} & \multicolumn{2}{c}{Filtered hybrid} \\
    \cmidrule(lr){2-3}\cmidrule(lr){4-5}\cmidrule(lr){6-7}\cmidrule(lr){8-9}
    System & nDCG@10 & R@100 & nDCG@10 & R@100 & nDCG@10 & R@100 & nDCG@10 & R@100 \\
    \midrule
    Reference (exact BM25 / exact cosine) & .308 & .530 & .350 & .609 & .361 & .621 & .342 & .588 \\
        Pinot lake-native (global BM25 statistics) & .312 & .533 & .348 & .603 & .361 & .622 & .345$^\dagger$ & .595$^\dagger$ \\
    \quad with per-segment IDF & .271 & .506 & & & .341$^\dagger$ & .612$^\dagger$ & & \\
    Elasticsearch 9.5.5 (1,000 candidates) & .312 & .533 & .339 & .588 & .354 & .620 & -- & -- \\
    OpenSearch 3.9.0 (ef\_search 1,000) & .312 & .533 & .342 & .593 & .356 & .621 & -- & -- \\
        \bottomrule
    \multicolumn{9}{p{0.92\textwidth}}{\footnotesize $^\dagger$On a fixed 2,000-query sample (seed 2026). On the same sample the exact filtered reference reaches .345 / .595, and lake-native hybrid with global statistics .360 / .623.}
  \end{tabular}
\end{table*}
Table~\ref{tab:quality} reports quality. Hybrid retrieval improves on both single modes in the exact reference
(nDCG@10 .361 vs.\ .308 BM25 and .350 vector), matching the complementarity reported on ESCI. Corpus-wide statistics matter: with per-segment IDF, lake-native BM25 falls to .271 nDCG@10; with the two-step
statistics protocol it reaches .312, within .004 of the exact reference. HNSW costs .003 nDCG@10 against exact search. BM25 is identical across Pinot, Elasticsearch and OpenSearch (.312, $p{=}0.49$), as expected from the same
analyzer and corpus statistics. Pinot's vector and hybrid runs are slightly higher (hybrid +.005 to +.007, $p<10^{-4}$;
+.013 to +.016 when the engines use 200 candidates) because Pinot ranks a larger candidate budget, the top 100 of
each file's graph (3,300 in all), against the engines' 1,000; this reflects the budget, not a better index.
With the broad brand filter, lake-native filtered hybrid and the exact filtered reference agree to the third decimal on
the 2,000-query sample (.345 / .595). Global statistics also carry into fusion: on the same sample, hybrid with per-segment IDF
scores .341 against .360 with global statistics.

\subsection{Ablations}\label{sec:abl}
\textbf{Corpus statistics.} Global statistics improve lake-native BM25 by +.040 nDCG@10 ($p<10^{-4}$); they win on
10,034 queries and lose on 3,685, with the rest tied. This is the one place where storing a table as many
independent files changes relevance, and Eq.~\eqref{eq:bm25} with summed statistics removes the effect. Rewriting
the same rows sorted by brand (a skewed layout) moves per-file BM25 from .272 to .293 nDCG@10 on the 2,000-query
sample, while global statistics give .314 under both layouts: with summed statistics, relevance no longer depends on
how the lake happens to cut its files.

\textbf{Fusion in SQL.} In-engine RRF matches client-side fusion (.3610 vs.\ .3605 nDCG@10).

\subsection{Latency and throughput}\label{sec:scale}
\begin{table}[t]
  \centering
  \caption{277M products (0.8\,TB, 2,052 files sorted by brand) on one i4i.4xlarge (16 vCPU, 128\,GB, local NVMe;
  indexes memory-mapped). Latency in ms. 1,000 queries (500 for the one-client vector run), each scoped to the most
  common brand among its relevant products; 32-client rows are the median of 20--40 passes after warm-up. BM25 and
  hybrid use per-snapshot corpus statistics (Section~\ref{sec:insights}; SQL RRF row with an earlier table covering
  95.4\% of queries).}
  \label{tab:scale}
  \setlength{\tabcolsep}{2.5pt}
  \begin{tabular}{lcccc}
    \toprule
    & 1 client & \multicolumn{2}{c}{32 clients} & Files \\
    \cmidrule(lr){2-2}\cmidrule(lr){3-4}
    Mode & p50/p99 & QPS & p50/p99 & searched \\
    \midrule
    Brand vector & 7.5/9.5 & 1,103 & 27/53 & 1 \\
    Brand BM25 & 6.5/47 & 635 & 40/176 & 1 \\
    Brand hybrid, client RRF & 8.9/48 & 375 & 75/246 & 1 \\
    Brand hybrid, SQL RRF & 33/74 & 122 & 252/389 & 1 \\
    \bottomrule
  \end{tabular}
\end{table}
Latency and throughput are measured on a lake of 277M products: replica 0 is the ESCI catalog and
each of 227 further replicas gets new product ids and embeddings perturbed by Gaussian noise (cosine 0.95 to the
original), written as 2,052 files of about 400\,MB and sorted by brand, as a lake clustered on a common filter column
would be. A brand-scoped query skips all but one of the 2,052 files from their statistics and searches that one,
re-applying the filter inside it, so a sorted rather than partitioned layout is safe~\cite{lakepann}. One node then
sustains 1,103 vector searches per second at 53\,ms p99, 635 BM25 searches and 375 hybrid searches fused by the
application (Table~\ref{tab:scale}); fusing in SQL runs through the multi-stage engine and adds about 25\,ms per query at one client.
Throughput depends on the working set: the same load spread over 21,000 queries and their brands gives 318 vector
searches per second, and a brand touched for the first time answers in 237\,ms (p50) while its pages load. A query
without a selective filter searches every file's graph, the case this design does not target. Servers share no
per-query state, so throughput should grow with the number of servers; a multi-server run is future work. A commit of 10,000 products to
this lake is searchable in 4.3\,s (median of three: 3.7, 4.3, 4.9\,s).

\subsection{Onboarding and storage}\label{sec:cost}
\begin{table*}[t]
  \centering
  \caption{Onboarding and storage for the same 1,215,854 products (EC2, each system alone). Extra bytes are
  what each system stores beyond the 4.38\,GB Iceberg table it was loaded from.}
  \label{tab:cost}
  \setlength{\tabcolsep}{5pt}
  \begin{tabular}{lcccccc}
    \toprule
    System & Load path & Time to queryable & Text index & Vector index & Row data & Extra bytes \\
    \midrule
    Pinot lake-native & register 33 files (3.4\,s), index at load & 149\,s & 0.27\,GB & 3.78\,GB & 0 & 4.06\,GB \\
    Elasticsearch 9.5.5 & bulk load (4 clients), refresh & 447\,s & 0.24\,GB & 3.78\,GB & 0$^\ast$ & 4.10\,GB \\
    OpenSearch 3.9.0 & bulk load (4 clients), refresh & 435\,s & 0.50\,GB & 3.78\,GB & 0$^\ast$ & 4.29\,GB \\
    \bottomrule
    \multicolumn{7}{l}{\footnotesize $^\ast$\texttt{\_source} disabled for a like-for-like comparison; keeping it stores the rows again.}
  \end{tabular}
\end{table*}

Registering the 33 files takes 3.4\,s; each segment then builds its text and HNSW indexes as it loads; the first
segment answers after 31\,s and the table after 149\,s, while a copy-based engine must first
read every row out of the lake, write it again and force-merge (Table~\ref{tab:cost}). Vector indexes dominate storage
everywhere (3.78\,GB), so the saving is the rows copied out of the lake and the pipeline that keeps them in sync.

\subsection{Infrastructure cost}\label{sec:infra}
\begin{table}[t]
  \centering
  \caption{Resources for a copy-based search engine and for Pinot on the lake, for two replicas, at 1.2M and 277M
  products (0.8\,TB); relative cost at low load from list prices and measured throughput, copy-based $=1$.}
  \label{tab:infra}
  \setlength{\tabcolsep}{3pt}
  \begin{tabular}{lcccc}
    \toprule
    & \multicolumn{2}{c}{1.2M products} & \multicolumn{2}{c}{277M (0.8\,TB)} \\
    \cmidrule(lr){2-3}\cmidrule(lr){4-5}
    & Copy & Pinot & Copy & Pinot \\
    \midrule
    Rows copied out of the lake & 8.8\,GB & 0 & 1.6\,TB & 0 \\
    Indexes & 8.2\,GB & 8.1\,GB & $\approx$1.9\,TB$^\ast$ & 2.2\,TB \\
    Sync pipelines & 1 & 0 & 1 & 0 \\
    Full build, one node & 707\,s & 149\,s & $\approx$45\,h$^\ast$ & 2.7\,h \\
    Nodes & 2 & 2 & 2$^\ddagger$ & 2$^\ddagger$ \\
    Relative cost & 1.00 & 0.91 & 1.00 & 0.98 \\
    \bottomrule
    \multicolumn{5}{p{0.95\columnwidth}}{\footnotesize $^\ast$Estimated from the 1.2M-product measurements, scaled by
    catalog size. $^\ddagger$One i4i.4xlarge per replica; Section~\ref{sec:scale} measures one.}
  \end{tabular}
\end{table}
A copy-based stack copies the rows out of the lake onto each replica (the \texttt{\_source} of a production index),
runs a sync pipeline, and rebuilds by bulk loading and force-merging; Pinot keeps only its indexes (Table~\ref{tab:infra}; on-demand
prices, pipeline as one small instance). At 1.2M products both need two nodes at low load, so Pinot costs 0.91 of the
copy-based stack, and 0.46 below 47 QPS when search joins an existing lake analytics cluster (one node per 47 hybrid QPS, the
measured single-node peak at this size, and no replicas of its own); it stays cheaper up to about 90 hybrid QPS. At
277M products, with two i4i.4xlarge replicas for either stack, the lake path avoids copying 1.6\,TB of rows and
running the pipeline, builds all its indexes in 2.7\,h on one node against an estimated 45\,h, and costs 0.98 of the
copy-based stack, since node count now dominates the bill. The comparison omits the engineering time a pipeline takes to run.

\subsection{Freshness}\label{sec:fresh}
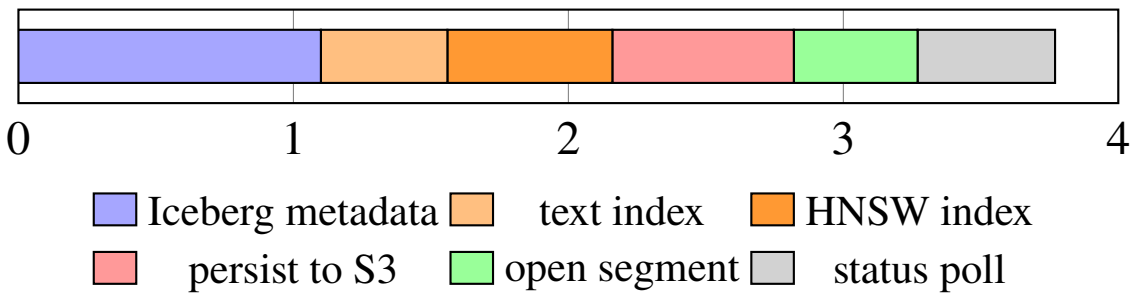
\begin{figure}[t]
  \centering
  \begin{tikzpicture}
    \begin{axis}[xbar stacked, width=\columnwidth, height=2.2cm, bar width=10pt, xmin=0, xmax=4,
        ytick=\empty, enlarge y limits=0.5,
        legend style={font=\footnotesize, legend columns=3, at={(0.5,-0.75)}, anchor=north, draw=none},
        xlabel style={font=\small}, ticklabel style={font=\small}]
      \addplot[fill=blue!35] coordinates {(1.1,0)};
      \addplot[fill=orange!50] coordinates {(0.46,0)};
      \addplot[fill=orange!80] coordinates {(0.6,0)};
      \addplot[fill=red!40] coordinates {(0.66,0)};
      \addplot[fill=green!40] coordinates {(0.45,0)};
      \addplot[fill=gray!35] coordinates {(0.5,0)};
      \legend{Iceberg metadata, text index, HNSW index, persist to S3, open segment, status poll}
    \end{axis}
  \end{tikzpicture}
  \caption{Where the 3.8\,s from a lake commit to searchable goes on EC2, in seconds (one run; 10,000 new products in one file).
  Index building is under a third of it.}
  \label{fig:fresh}
\end{figure}
We commit 10,000 new products to the Iceberg table as one Parquet file and measure the time until all are searchable
(50 sampled unique tokens return their products and the row count matches; 1\,s polling, three runs). For Elasticsearch and OpenSearch we hand the copy
pipeline the change set for free (no change-data-capture lag), bulk-index it and refresh. On EC2, commit-to-searchable
(median of three) is 3.8\,s for Pinot lake-native and 3.1\,s and 3.6\,s for Elasticsearch and OpenSearch, whose numbers
exclude any pipeline lag. Figure~\ref{fig:fresh} breaks the 3.8\,s down: the new file's indexes take 1.1\,s, while
reading Iceberg metadata from S3 (1.1\,s) and persisting the indexes before the segment goes online (0.7\,s) together
take longer; the latter can move off the critical path. Every query sees whole snapshots, never a half-applied batch.

\section{Design Insights}\label{sec:insights}
\textbf{Per-file indexes trade fan-out for incremental freshness.} A lake commit indexes only its new file (1.1\,s on
EC2) and nothing is ever merged, while an engine with one graph per index force-merges after a bulk load (260\,s and
231\,s). File size is the knob between how much a commit re-indexes and how many graphs a query searches.

\textbf{Freshness is bounded by metadata and persistence, not indexing.} Building a new file's indexes is under a third
of the time from commit to searchable (Figure~\ref{fig:fresh}); the rest is catalog metadata, persistence and
coordination, which is where the next second will come from.

\textbf{Indexes on a remote tier need zero-copy reads.} Serving HNSW reads through Lucene's buffered input copied every
page it touched; handing Lucene the index buffer directly raised vector throughput by 34\% on EC2 with identical results.

\textbf{Pruning keeps the working set in memory.} At 277M products the indexes (1.1\,TB) are over eight times the node's memory,
yet a brand-scoped query touches only the files that survive pruning, whose pages stay in the page cache once warm
(237\,ms on first touch, 7.5\,ms warm). Pruning must also see files where the filter column is entirely null: their
min/max statistics are empty, so the pruner kept all 122 brand-less files and every brand query searched 123 graphs.
Reading the null count recorded in the segment metadata and skipping such files cut that to one graph and raised
vector throughput 25 times, from 30 to 760 searches per second before further tuning. Sorting or clustering the Iceberg table on the columns queries filter by (brand,
category, seller) turns a search over the lake into a search over a few files.

\textbf{Corpus statistics are per snapshot, not per query.} Gathering BM25 statistics touches every file's text
index, so at 2,052 files it cost more than the scored search itself (26 vs.\ 6.5\,ms). The statistics change only
with the snapshot, so we materialize them for the whole vocabulary (658,669 terms, 19.5\,min per snapshot) and serve
them from memory; 97.6\% of queries need no statistics request, rankings are identical to live statistics on all 179
queries we checked, and BM25 throughput rose from 60 to 635 searches per second.

\textbf{Search and analytics on one copy.} Retrieval is SQL, so results can be faceted or aggregated in the same
query (e.g.\ \texttt{GROUP BY brand} over the hybrid top-$k$), over lake and stream alike.

\section{Conclusion}
Search applications do not need a copy of the catalog. With BM25 over corpus-wide statistics, ANN with filter pushdown
and RRF in SQL, an engine that indexes the lake's own files matches exact retrieval on quality, becomes searchable
minutes after files are registered and seconds after a commit, and on one node serves selective searches over 277M
products in milliseconds, at more than 1,000 vector searches per second. The lake stays the single system of record, streams can join
it behind one table, and the copy, the sync pipeline and the work of running them go away. The scoring and statistics
functions will be proposed to the Apache Pinot community.

\section*{Artifact Availability}
Benchmark scripts, table configurations and per-query results are available from the authors on request.

\bibliographystyle{IEEEtran}
\bibliography{refs}
\end{document}